\documentclass[pre,twocolumn,unsortedaddress,floatfix,amssymb,aps,10pt]{revtex4-2}
\usepackage{amsmath}
\usepackage{graphicx}
\usepackage{dcolumn}
\usepackage{bm}

\usepackage[utf8]{inputenc}
\usepackage[T1]{fontenc}
\usepackage{mathptmx}
\usepackage{etoolbox}

\makeatletter
\def\@email#1#2{%
 \endgroup
 \patchcmd{\titleblock@produce}
  {\frontmatter@RRAPformat}
  {\frontmatter@RRAPformat{\produce@RRAP{*#1\href{mailto:#2}{#2}}}\frontmatter@RRAPformat}
  {}{}
}%
\makeatother
\begin{document}

\preprint{AIP/123-QED}

\title{Dimensional criterion for forecasting nonlinear systems by reservoir computing}
\author{Pauliina T. K\"arkk\"ainen}
\author{Riku P. Linna*}
\email{Corresponding author: riku.linna@aalto.fi}
\affiliation{
  Department of Computer Science,
  Aalto University, P.O. Box 15400, FI-00076 Aalto, Finland
 }


\begin{abstract}
Reservoir computers (RC) have proven useful as surrogate models in forecasting and replicating systems of chaotic dynamics. The quality of surrogate models based on RCs is crucially dependent on  their optimal implementation that involves selecting optimal reservoir topology and hyperparameters. By systematically applying Bayesian hyperparameter optimization and using ensembles of reservoirs of various topology we show that connectednes of reservoirs is of significance only in forecasting and replication of chaotic system of sufficient complexity. By applying RCs of different topology in forecasting and replicating the Lorenz system, a coupled Wilson-Cowan system, and the Kuramoto-Sivashinsky system, we show that simple reservoirs of unconnected nodes (RUN) outperform reservoirs of connected nodes for target systems whose estimated fractal dimension dimension is $d \lesssim 5.5$ and that linked reservoirs are better for systems with $d > 5.5$. This finding is highly important for evaluation of reservoir computing methods and on selecting a method for prediction of signals measured on nonlinear systems.
\end{abstract}

\pacs{}

\maketitle 

\section{Introduction}



Reservoir computers (RC) have been found useful in replication and forecasting of chaotic systems~\cite{ lukosevicius09, jaeger12, pathak17, pathak_prl18}. Originally, reservoir computing was defined as a supervised learning method where the reservoir is a fixed recurrent echo-state network~\cite{jaeger01, maas02, jaeger12}. Topology of the recurrent neural network (RNN) is considered one of the central aspects regarding performance of RCs. In search of an optimal topology, random Erd\"os-R\'enyi (ER) networks are typically used as reference. The performance of different reservoir networks have been compared to that of ER networks for reservoirs as large as $10 000$ nodes. However, no clear preferential topology has been found for reasonably densely connected RNNs~\cite{lukosevicius09}.

Sparsely connected RCs were considered from the very beginning. Jaeger et al pointed out that sparse interconnectivity within the reservoir of a RNN could be beneficial in that it lets the reservoir decompose into many loosely coupled subsystems, establishing a richly structured reservoir of excitable dynamics~\cite{jaeger12}. Partly supportive of Jaeger's intuition, judiciously selected low-connectivity reservoirs were recently found to perform as well as densely connected reservoirs in forecasting chaotic systems by Griffith et al.~\cite{griffith19}. They also identified the importance of optimizing hyperparameters and applied a rather judicious optimization procedure.

Despite some promising systematic approaches, see e.g.~\cite{maas02,dambre12, lu18, smith22, bollt21}, a comprehensive theoretical understanding of reservoir computing is still missing. Due to the insufficient theoretical framework, optimization of RCs needs to be done by experimentation. It is fair to say that generally evaluations of RCs for predicting systems of chaotic dynamics have been made using a very low number of RCs, or even a single RC, which has led to some misleading conclusions. This deficiency was first addressed by Haluszczynski et al.~\cite{haluszczynski19}.

Since reservoirs of RCs were originally defined as RNNs, introduction of the echo state property (ESP) of the reservoir network was essential~\cite{jaeger01,jaeger12, yildiz12}. Unlike feed forward networks, RNNs are nonlinear dynamical systems that may exhibit instability and bifurcations. Requiring the reservoir to have ESP aids in designing a RNN, whose outcome converges toward the data to be learned and predicted. However, there exist RC implementations where linked reservoirs are not RNNs. In this paper the reservoir of unconnected nodes (RUN) is found well suited for forecasting low-dimensional systems of chaotic dynamics. RUN is a feed forward network and as such can fulfill ESP only via leakage in the RC. Despite this, we find that several optimized RUNs that perform best in forecasting low-dimensional chaotic systems have no leakage and thus do not fulfill ESP. For high-dimensional systems exhibiting spatiotemporal chaos, RUNs without ESP are found to always perform better than RUNs that posses ESP.

In keeping with our findings on forecasting and replication of low-dimensional chaotic systems, a simplified RC, where identity function is used as internal activation function to facilitate analysis, was recently found to perform surprisingly well for low-dimensional chaotic systems~\cite{bollt21}. Moreover, a simple regularized regression method was found to yield better prediction than linked RCs for the Lorenz system~\cite{pyle21}. In this paper we show that simple methods, such as an unconnected RC, perform better than RCs only for sufficiently low-dimensional systems. For more complex, higher-dimensional systems, the dynamical aspect of linked RCs becomes important, and such RCs outperform simpler methods. The requirement of the system complexity turns out to be higher than traditionally expected, which is important to take into account in future applications of and research on using RCs for forecasting and replication of nonlinear systems. 

In what follows, we use ensembles of carefully optimized reservoirs of different topology to determine optimal RCs for forecasting and replication of the Lorenz, coupled Wilson-Cowan, and Kuramoto-Sivashinsky systems. We show that, contrary to a general assumption, the dynamics of the system under study being chaotic is not a sufficient condition for using reservoir computing instead of more conventional methods like delay embedding~\cite{kantz04}. We show that forecasting and replication benefits from reservoir computing using linked reservoirs only when the attractor of the chaotic system has fractal dimension $d_c \approx 5.5$ or greater. For dimension smaller than $d_c$, RC with a reservoir of unconnected nodes (RUN) replicates and forecasts better than any RC with a reservoir including connected nodes. 


The original definition of RCs having an RNN as a reservoir excludes even linked reservoirs without loops. In order to avoid unnecessary complications we will use the term reservoir computer (RC) for all the supervised learning methods that include a reservoir, regardless of its topology.

The paper is organized as follows. Section~\ref{RCs} describes the implementation of the used RCs and how they are evaluated. In Section~\ref{dynamical systems} the dynamical systems used in forecasting and prediction are depicted. Results are given in Section~\ref{results}. RCs in connection with low-dimensional and high-dimensional systems are studied in Sections \ref{low-dim} and \ref{high-dim}, respectively. In Section~\ref{conclusion} we explain our findings by writing down equations for unconnected and linked RCs in the relevant frameworks. We show that the delayed signals that are present only in linked RCs, making them dynamical systems, is the only difference between unconnected and linked RCs that explains the found dimensional criterion.

\section{Reservoir Computer Implementation and Evaluation}
\label{RCs}

\begin{figure}
  \centering
    \includegraphics[width=0.45\textwidth]{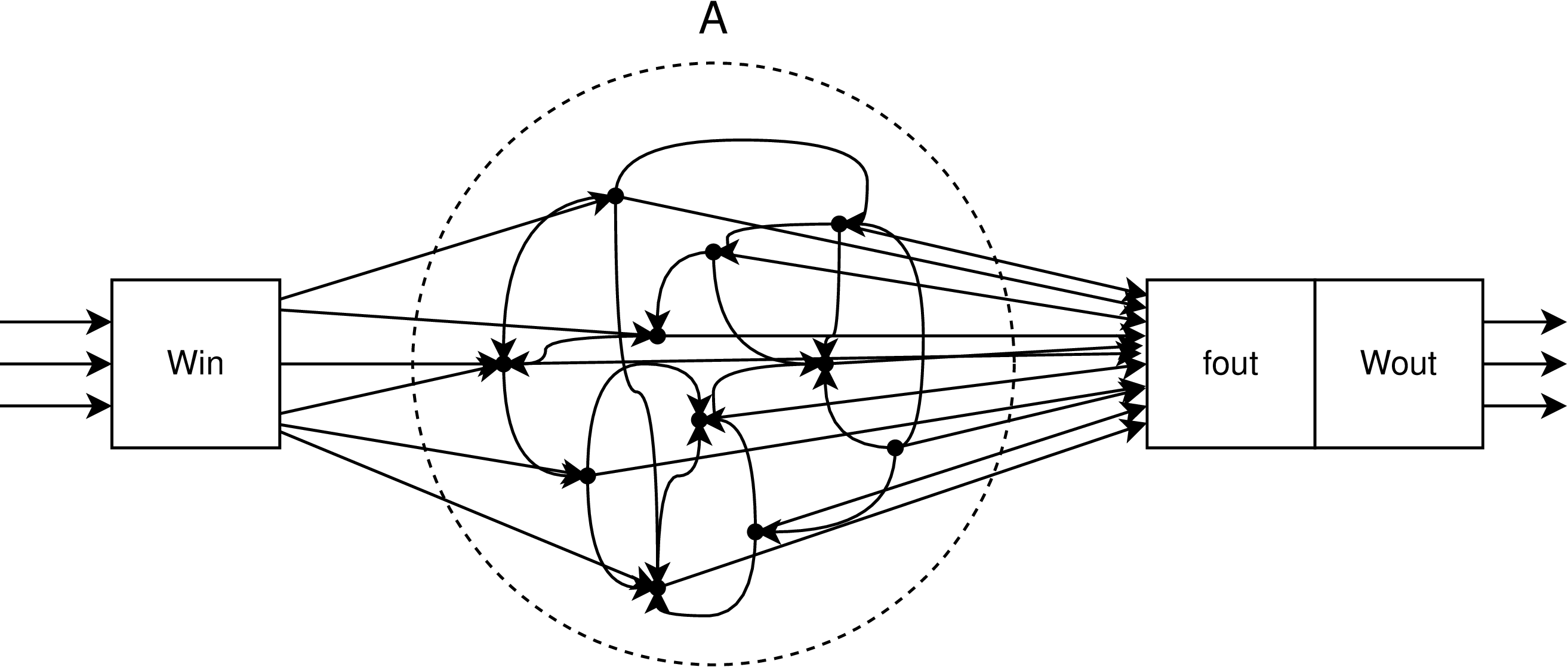}
  \caption{Schematic depiction of RC in the training phase. In the prediction phase the output of the reservoir is fed back to the input and the reservoir runs autonomously.}
  \label{fig:reservoirdiagram}
\end{figure}

A reservoir computer (RC) comprises an input layer, an artificial neural network, and an output layer, see Fig.~\ref{fig:reservoirdiagram}. Network connections are weighted and directed. We use the number of connections to a node, in-degree $k$, as the measure of network connectivity.  The connections from the input to the $D_r$ nodes of the network are determined by the input matrix $\mathbf{W}_{in}$. Below, the differential equation and its numerical form for time evolution of the reservoir state $\mathbf{r}(t)$ are given on the first and the second row, respectively.
%
\begin{eqnarray}
\label{reservoir_dynamics}
\mathbf{\dot{r}}(t) &=& -\gamma \mathbf{r}(t) + \gamma \tanh{[\mathbf{A}\mathbf{r}(t) + \mathbf{W}_{in}\mathbf{B}(t)]} \\
\mathbf{{r}}(t+\Delta t) &=& (1-\beta) \mathbf{r}(t) +\beta\tanh{[\mathbf{A}\mathbf{r}(t) + \mathbf{W}_{in} \mathbf{B}(t)]}. \nonumber
\end{eqnarray}
Here, $\mathbf{A}$ and $\mathbf{r}(t)$ are the reservoir's adjacency and state matrices, respectively. $\gamma$  defines the natural rate, or the inverse time scale, of the reservoir dynamics. In the numerical form $\Delta t$ is the time step and $\beta = \gamma \Delta t$ the leakage parameter~\cite{lu17}.

In training RC the driving signal $\mathbf{u}(t)$ from the dynamical system is fed to the reservoir network, $\mathbf{B}(t) = \mathbf{u}(t)$. The input matrix $\mathbf{W}_{in}$ and the topology of the reservoir via $\mathbf{A}$ are determined before training and remain fixed after this. Input connections are first established with a given probability $p_{in}$, after which the strength for each existing connection is taken as a number chosen randomly from a standard normal distribution. The elements of $\mathbf{A}$ and $\mathbf{W}_{in}$ are scaled such that the given spectral radii $\rho_r$ and $\rho_{in}$ are obtained.

To exclude the initial transient, half of $\mathbf{r}(t)$ from the beginning is discarded. The output layer then transforms the remaining reservoir's output $\mathbf{r}^r$ to $\mathbf{W}_{out} \mathbf{\tilde{r}}(t)$. Transformation $\mathbf{\tilde{r}}(t) = \mathbf{f}_{out}\mathbf{r}^r(t)$ removes unwanted symmetries in the reservoir that in the combined system of the reservoir and the dynamical system may deteriorate prediction~\cite{pathak17}. The symmetry-breaking form of $\mathbf{f}_{out}$ transforms the reservoir node values as $\tilde{r}_i(t) = r_i(t)$ for $i \le N/2$ and $r_i(t)^2$ for $i > N/2$, where $N$ is the number of nodes in the reservoir.

In the next training stage $\mathbf{W}_{out}$ is adjusted such that  $\mathbf{W}_{out} \mathbf{\tilde{r}}(t)$ approximates the output $\mathbf{u}(t)$ of the dynamical system that is known for a time interval $t \in [0, T_{train}]$, where $T_{train}$ is the training time. This is done by minimizing the quantity $\sum_{t=0}^{T_{train}} |\mathbf{u}(t) - \mathbf{W}_{out} \mathbf{\tilde{r}}(t)|^2 + \mu||\mathbf{W}_{out}||^2$, where $\mu$ is the ridge regression parameter. After this  $\mathbf{W}_{out}$ remains fixed. Finally, forecasting is done by running the trained RC autonomously. At this stage $\mathbf{B}(t) = \mathbf{W}_{out} \mathbf{\tilde{r}}(t)$ in Eq.(\ref{reservoir_dynamics}) and the output $\mathbf{W}_{out} \mathbf{\tilde{r}}(t)$ is continuously fed back in the input.

In the numerical form of Eq.(\ref{reservoir_dynamics}) we use $\Delta t = 0.01,\ 0.3$, and $0.25$ for the Lorenz, coupled Wilson-Cowan (cW-C), and Kuramoto-Sivashinsky (K-S) systems, respectively. For all systems time $t$ is given in the same "bare" units, i.e. $t = n \times \Delta t$, where $n$ is number of timesteps of length $\Delta t$ for each system. This way times for all systems are directly comparable. $\beta \in |0, 1]$  determines how large a portion of the past state of the reservoir is directly repeated in the present state. For $\beta = 1$ there is no direct feedback and the present state is determined only through neural-network-type evolution. For $\beta < 1$, RC is a leaky integrator~\cite{yildiz12} and has ESP, regardless of the reservoir topology.

Hyperparameter values are found by Bayesian optimization. Optimal values are found for six parameters~\cite{griffith19}: spectral radius $\rho_r$, probability of connecting an element of $\mathbf{W}_{in}$ to RC $p_{in}$, spectral radius of $\mathbf{W}_{in}$ $\rho_{in}$, leakage rate $\beta$, regression parameter $\mu$, and in-degree $k$. This hyperparemeter set is conceptually complete: RC can be in principle optimized through them.

During forecasting we evaluate the performance of RC over time $T_{eval}$ by the short time prediction error averaged over $P$ start times 
\begin{equation}
\epsilon = \Bigg\{\frac{1}{P}\sum_{i=1}^P \epsilon_i^2\Bigg\}^{1/2},
\label{rmserror}
\end{equation}
where $\epsilon_i^2 = \frac{\Delta t}{T_{eval}}\sum_{t=t_i}^{t_i+T_{eval}}|\mathbf{u}(t)-\mathbf{W}_{out}\mathbf{\tilde{r}}(t)|^2$. We use $T_{eval} = 1/\lambda_1$, where $\lambda_1$ is the largest Lyapunov exponent of the respective system in the chaotic regime, for the Lorenz and cW-C systems in all regimes. $T_{eval} = 1/\lambda_1$ of the  K-S system with $L = 35$ is used for all K-S systems. Before each start time $t_i$, the reservoir is run with input for the time from $t_i-\xi$ to $t_i$ in order to synchronize the state of the reservoir to the state of the system. We use $P=50$ and synchronization times $\xi = 11$, $330$ and $10$ for the Lorenz, cW-C, and K-S systems, respectively. In a chaotic system the $P$ start times define points in the attractor, so $\epsilon$ is calculated as an average over separate trajectories. We both minimize $\epsilon$ in determining the optimal hyperparameter values and use it as a figure of merit for evaluating RC performance.

A figure of merit most directly related to signal prediction is the valid time $T_v$, defined as the elapsed time before the normalized error $E(t) = ||\mathbf{u}(t) - \mathbf{W}_{out}\mathbf{\tilde{r}}(t)||/\langle||\mathbf{u}(t)||^2\rangle^{1/2}$ exceeds some value $f \in [0, 1]$. Here, $f = 0.4$ and $||\cdot||$ denotes $\textrm{L}_2$- norm. $T_v$ is averaged over $20$ start times $t_i$.

Climate replication, that is, the capacity of RCs to reproduce ergodic properties of chaotic systems, is evaluated by measurements of Lyapunov exponents $\lambda_i$ and the fractal dimension $d_f$ of the attractor. $\lambda_i$ of RCs are computed as in~\cite{pathak17}. $d_f$ is estimated numerically by computing correlation dimension $d_c$ using Grassberger-Procaccia algorithm~\cite{grassberger_a83, grassberger_b83}  and via measured $\lambda_i$ by Kaplan-Yorke dimension
\begin{equation}
d_{KY} = j + \sum_{k = 1}^j \frac{\lambda_k}{\vert \lambda_{j+1} \vert},
\label{Kaplan-Yorke}
\end{equation}
where j is the largest integer for which the cumulative sum of the Lyapunov exponents is positive~\cite{frederickson, carroll20b}.

\section{Dynamical systems}
\label{dynamical systems}

We use three dynamical systems for optimization and evaluation of RCs. These systems are used to generate simulated data $\mathbf{u}(t)$ extending over time $t \in [0, T]$.

\subsection{The Lorenz system}
\label{lorenz}

The Lorenz system~\cite{lorenz63} is defined as
\begin{equation}
    \dot{x} =  \sigma(y-x),\ \dot{y} = rx - y -xz, \ \dot{z} = xy - bz,
    \label{lorenz}
\end{equation}
where $\dot{x} = dx(t)/dt$. Unless otherwise noted, this system is used in the chaotic regime with the parameter values $\sigma = 10$, $b = 8/3$, and $r = 28$, originally used by Lorenz. This is the most generally used test bench for RCs. We also evaluate RCs as surrogate models for the Lorenz system in intermittently chaotic  ($r = 100$) and periodic regimes ($r = 150$).

\subsection{The system of coupled Wilson-Cowan models}
\label{wilson-cowan}

The dynamical system of two reciprocally connected Wilson-Cowan models is defined as~\cite{maruyama14}
\begin{align}
\tau_e \dot{E}_l &= -E_l + (k_e -E_l)\cdot  S_e(c_1E_l-c_2I_l+P_l+\alpha E_m), \nonumber\\
\tau_i \dot{I}_l &= -I_l + (k_i -I_l)\cdot S_i(c_3E_l-c_4I_l),
\end{align}
where the indices are $l = 1$ and $m = 2$ for the first and $l = 2$ and $m = 1$ for the second W-C model. $\tau_e$ and $\tau_i$ are time constants of excitatory and inhibitory neurons, respectively, $E_l(t)$ and $I_l(t)$ are average activities of groups of these neurons at time $t$, $c_1, c_2, c_3$, and $c_4$ are coupling strengths between and within neural groups, and $k_e$ and $k_i$ are constants. The coupling strength of the two Wilson-Cowan models is defined by $\alpha$. $S_e$ and $S_i$ are sigmoidal functions $S_e(x) \equiv \{1+ \textrm{exp}(-a_e(x-\theta_e))\}^{-1} - \{1+\textrm{exp}(a_e\theta_e)\}^{-1}$ and $S_i(x) \equiv \{1+\textrm{exp}(-a_i(x-\theta_i))\}^{-1} - \{1+\textrm{exp}(a_i\theta_i)\}^{-1}$. Parameter values $\theta_e=4.0$, $a_e = 1.3$, $\theta_i=3.7$, and $a_i = 2.0$, $P_1=1.09$, $P_2=1.06$, $\tau_e = \tau_i = 1.0$, $k_e=k_i=1.0$, $c_1=16.0$, $c_2=12.0$, $c_3= 15.0$, and $c_4=3.0$ were used.

RC forecasting and replication was evaluated for three different regimes: aperiodic (chaotic) ($\alpha = 1.3$), quasiperiodic ($\alpha = 1.9$), and periodic ($\alpha = 4$).


\subsection{The Kuramoto-Sivashinsky system}
\label{kuramoto-sivashinsky}

The one-dimensional Kuramoto-Sivashinsky equation for $y(x,t)$ is defined as
\begin{equation}
 y_t =  -yy_x - y_{xx} - y_{xxxx},
\label{K-S}
\end{equation}
where $y_x = \partial y(x,t)/\partial x$. $L$ is the domain size: $ 0 \leq x < L$. Periodic boundary conditions $y(x,t) = y(x+L, t)$ are applied. The term $-y_{xxxx}$ is stabilizing while $-y_{xx}$ drives the system to instability. Taking the Fourier transform of Eq.~(\ref{K-S}) results in $\dot{Y}_k(t) =  (k^2-k^4) Y_k(t) - \frac{ik}{2} \mathcal{F}\{y^2(t)\}$, where $Y_k(t) = \mathcal{F}\{y(t)\}$ is the Fourier transform of $y(t)$. It is seen that the system gets increasingly unstable as $k$ decreases, that is, the wavelength increases. $L$ acts as a bifurcation parameter: Increasing $L$ allows larger wavelengths and brings the system  that has stable traveling waves for $L \lesssim 13$ to spatiotemporal chaos~\cite{edson19}. Simulated data $\mathbf{u}(t) = [ y(L/Q, t), y(2L/Q, t), .. , y(L, t)]$ for training and evaluation of RCs is obtained by numerical integration of Eq.~(\ref{K-S}) on a one-dimensional equally spaced grid of size $Q = 64$.


\section{Results}
\label{results}

 \subsection{Low-dimensional dynamical systems}
 \label{low-dim}

 \subsubsection{Forecasting the standard chaotic Lorenz system}
 \label{topology}

We first verify our RC implementation by reproducing the result in~\cite{griffith19} suggesting thar reservoir topology affects RC performance on forecasting the standard chaotic Lorenz system. Identical ranges for hyperparameter optimizations, see the first row, labeled R, in Table~\ref{tab:ranges} and training time, $T_{train} = 100$, were used. To evaluate the optimization procedure consisting of $100$ iterations we ran it for $20$ times for each topology. $200$ reservoirs of size $D_r = 100$ were generated using median parameter values from the optimization. Random Erd\"os-R\'enyi (ER) networks, where $k$ ranges from $1$ to $5$, were simulated together with RCs with all possible singly-connected ($k = 1$) reservoir networks that include at most one cycle: a reservoir including a single cycle (ISC), a reservoir of single tree including all nodes (ST), a single cycle including all the nodes (SC), and a single line including all the nodes (SL).
 
\begin{figure}[ht]
  \centering
    \includegraphics[width=0.48\textwidth]{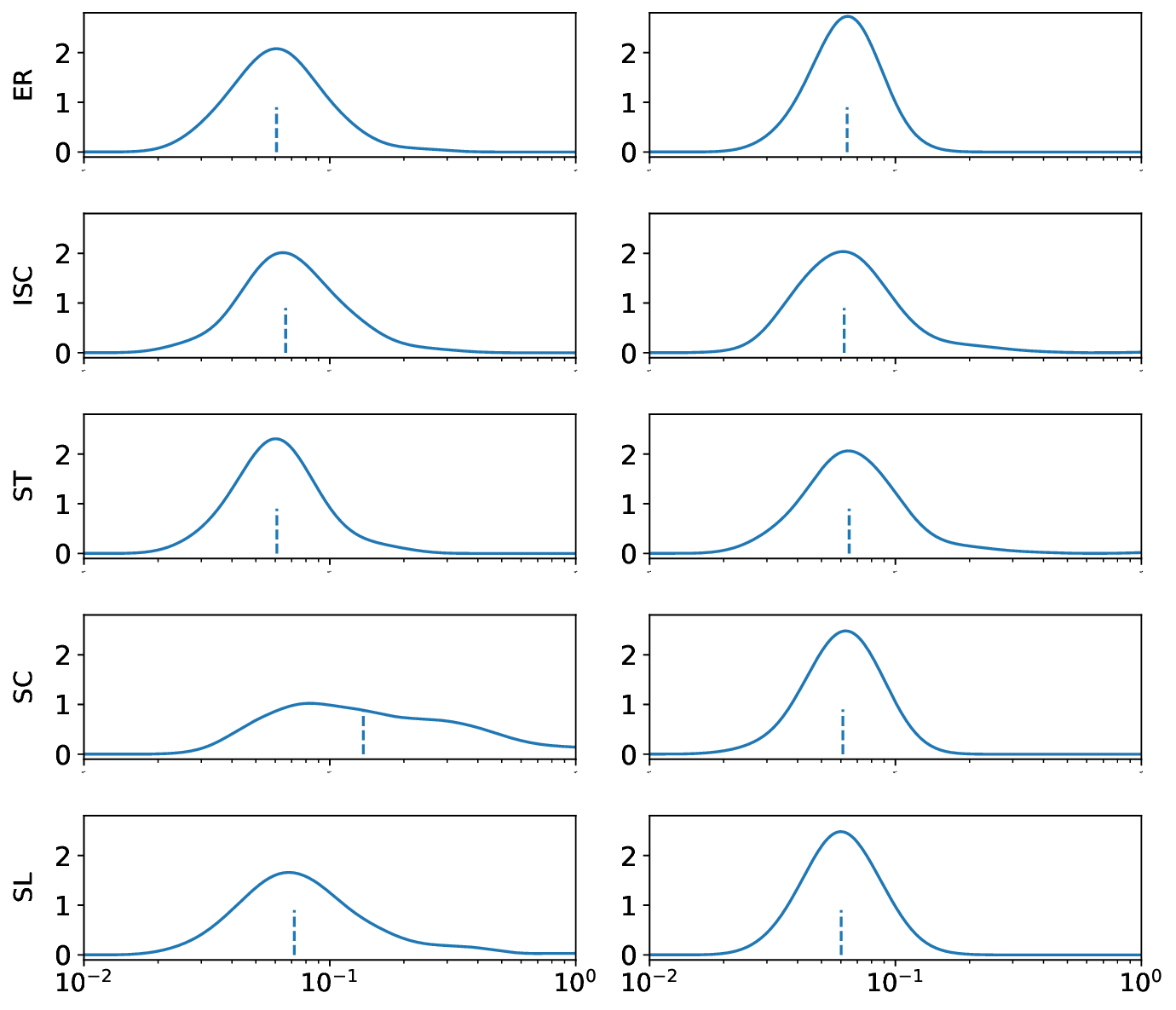}
  \caption{Distributions of $\epsilon$ from $200$ simulations, visualized as Gaussian kernel density estimations in $\log(\epsilon)$ with bandwidth $0.1$~\cite{scott92}. Reservoirs top down: ER, ISC, ST, SC and SL (see text for explanation). The first column: hyperparameters the same as in~\cite{griffith19} and obtained from one optimization. The second column: hyperparameters from $20$ runs of Bayesian optimization. Dashed lines show median values.}
  \label{fig:astpdist}
\end{figure}

 Fig.~\ref{fig:astpdist} shows distributions of short time prediction errors $\epsilon$, Eq.~(\ref{rmserror}), for $200$ realizations of each reservoir topology. The vertical lines show the median values  $\textrm{Med}(\epsilon)$. For the plots in the first column we used hyperparameter values that in~\cite{griffith19} were found optimal after one optimization. Our RC implementations give very similar results as in~\cite{griffith19}. Most notably, SC would seem to give the largest $\epsilon$.

A more thorough optimization within the same set of hyperparameter ranges, R in Table~\ref{tab:ranges}, changes the outcome. The second column of Fig.~\ref{fig:astpdist} shows $\epsilon$ when the hyperparameter values were obtained, not from a single optimization, but as medians out of $20$ optimizations and used for $200$ RCs of each topology. There are no discernible differences between distributions of $\epsilon$ for reservoirs of different topology.

It is noteworthy that ST and SL networks do not have ESP in the strict mathematical sense. They do not forget the present state asymptotically like an RNN, but in a finite number of steps. All reservoirs that have leakage ($\beta < 1)$, including RUN, posses the strict ESP. For all practical purposes, the criterion reservoir forgetting its state in a finite but large number of steps serves as ESP.

 
 \begin{table}
\centering
\begin{tabular}{c|c|c|c|c|c|c}
 & $\rho_r$ & $p_{in}$ & $\rho_{in}$ & $\beta$ & $\log{\mu}$ & $k$\\\hline
R & [0.3, 1.5] & [0.0, 1.0] & [0.3, 1.5] & [0.07, 0.11] & [-5, 5] & [1, 5]\\
A & [0.1, 1.5] & [0.1, 1.0] & [0.1, 1.5] & [0.05, 1.0] & [-5, 0] & [1, 5]\\
B & [0.0, 1.5] & [0.0, 1.0] & [0.0, 1.5] & [0.05, 1.0] & [-5, 0] & [1, 5]\\
\end{tabular}
\caption{\label{tab:ranges}Hyperparameter optimization ranges for ER reservoirs.}
\end{table}

Importantly, we also found that data representation has a clear impact on forecasting. In the above, the variance of the data $\mathbf{u}(t)$ was normalized to make comparison with~\cite{griffith19}. This normalization, commonly used in gradient-descent based learning, does not preserve the relative magnitudes of $u_i(t)$, where $i$ denotes dimensions. This could deteriorate supervised learning. Our experiments proved this to be the case. In our simulations of ER reservoirs, using data as is or normalizing the data with the difference of its maximum and minimum value over time $t \in [0, T_{train}]$ resulted in clearly smaller $\epsilon$ and longer $T_v$. As the best RC performance was obtained by the data normalization we apply it in what follows.


 
Prompted by the found optimal hyperparameter values lying close to the limits of the allowed ranges, see the Supplement, we use wider ranges in optimizing hyperparameters for ER, see Table~\ref{tab:ranges}.  Only linked reservoirs are allowed in A to keep optimized reservoirs ER. In order to see the resulting optimum reservoirs without this restriction, RUNs are allowed in B by including $\rho_r = 0$ . Optimization was run for $200$ iterations for A and $500$ iterations for the wider ranges of B. Using the longer $T_{train} = 1000$ was seen to slightly improve the RC performance. Optimal hyperparameter values, errors $\epsilon$, and valid times $T_v$ are shown in Table~\ref{tab:parameters_vs_data}.
\begin{table}
\centering
\begin{tabular}{|c|c|c|c|c|c|c|c|c|c|}
\hline
Range & $T_{train}$ & $\rho_r$ & $k$ & $p_{in}$ & $\rho_{in}$ & $\beta$ & $\log{\mu}$ & $\epsilon$ & $T_v$  \\\hline
A & 100 & 0.74 & 2 & 0.77 & 1.28 & 0.39 & -5 & 0.02 & 5.20\\
B & 100 & 1.01 & 1 & 0.64 & 1.31 & 0.35 & -5 & 0.02 & 5.15\\
A & 1000 & 0.95 & 2 & 0.54 & 1.50 & 0.55 & -5 & 0.03 & 5.28\\
B & 1000 & 0.00 & - & 0.89 & 1.50 & 0.62 & -5 & 0.02 & 6.11\\
\hline
\end{tabular}
\caption{\label{tab:parameters_vs_data}ER reservoirs of size  $D_r = 100$. Optimal hyperparameter values, errors $\epsilon$, and valid times $T_v$ obtained for $200$ RCs using sets A and B of hyperparameter ranges, see Table~\ref{tab:ranges}. On the last row the optimum value obtained for $k$ has no significance, since $\rho_r = 0$ means that $k = 0$.
} 
\end{table}

From the combined values of $\rho_r$ and $k$ reservoirs are seen to be sparse for B . Optimization resulted in RUN ($\rho_r = 0$) in $70 \%$ of the cases for $T_{train} = 1000$. We also made sure that the found optimal reservoirs of very low connectivity are not just optima within the chosen hyperparameter ranges by using the wider range of $k \in [1, 10]$ and several ranges for $\rho_r$. These optimizations resulted in minimal $k$ values. To summarize, optimization that allows high connectivity leads to very sparsely connected reservoirs and for sufficienlly long $T_{train}$ the optima of these sparsely connected reservoirs are RUNs. 

\begin{figure}
  \centering
    \includegraphics[width=0.48\textwidth]{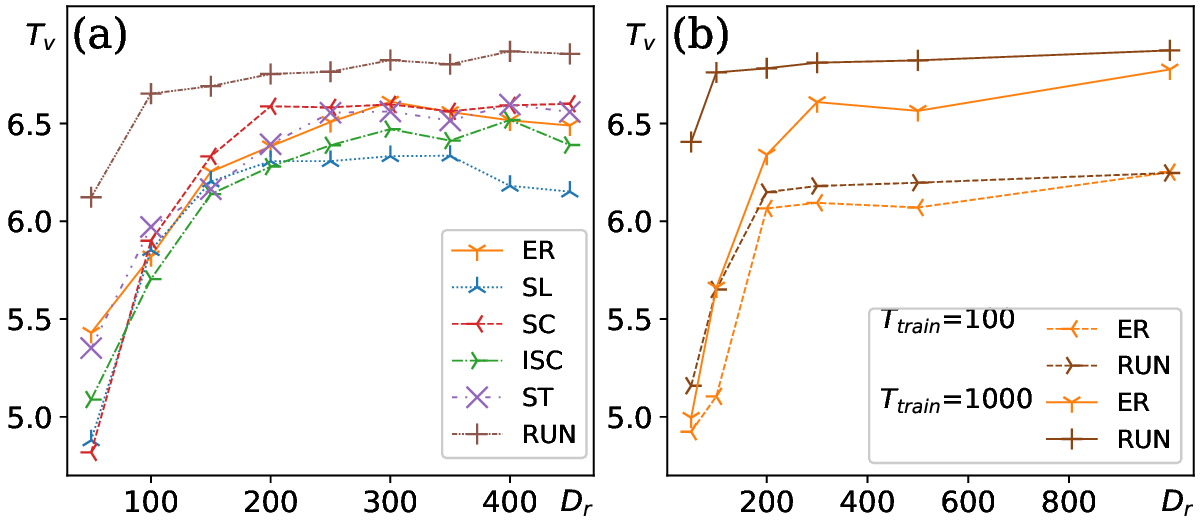}
  \caption{The standard chaotic Lorenz system. Valid time $T_v$ vs reservoir size $D_r$. a) All the five topologies. $200$ RCs of each size $D_r \in \{50,\ 100,\ 150, \ldots,\ 450\}$. Hyperparameters obtained from optimization of $20$ RCs of size $D_r^{O} = 300$. $T_{train} = 1000$ ($10^5$ steps of of size $\Delta t = 10^{-2}$).
  b) ER and RUN.$1000$ reservoirs of each size $D_r \in [50,100,200,300,500,1000]$. Hyperparameters obtained from optimization of $50$ RCs of each $D_r$. $T_{train} = 100$ and $1000$.
  }
  \label{fig:errundr}
\end{figure}

Fig.\ref{fig:errundr}(a) shows measured $T_v$ for reservoirs of different topology and size when hyperparameter values were obtained for reservoirs of size $D_r^{O} = 300$. The best performance for each topology is obtained for reservoirs of size $D_r \ge D_r^{O}$. In other words the performance does not improve by using reservoirs larger than for which the optimization was done. Qualitatively, results are the same for all $D_r$. RUN is seen to perform best for all $D_r$. ER is seen to perform better than the the singly-connected RCs, which makes it a valid reference in what follows.

We compared performances of ER and RUN for even larger $D_r$, see  Fig.~\ref{fig:errundr}~b). Here, optimizations within ranges A was done for each $D_r$ separately to eliminate the possible effect of having $D_r \ne D_r^{O}$. RUN outperforms ER for all $D_r$. Essentially the best performance for RUN is reached for $D_r \gtrsim 300$. Increasing $T_{train}$ is seen to improve the performance of both RCs.


In summary, optimization of RCs for the standard chaotic Lorenz system leads to sparsely connected reservoirs and RUNs. RUNS were verified to outperform all sparsely linked reservoirs in forecasting the standard chaotic Lorenz system.

\subsubsection{Forecasting the Lorenz and coupled Wilson-Cowan systems}

To make sure that our findings are not specific to the Lorenz system, we compare ER and RUN in forecasting the Lorenz and the coupled Wilson-Cowan  (cW-C) systems in different dynamical regimes. Table~\ref{tab:LWC} shows $\epsilon$ and $T_v$ measured for $100$ RCs of $D_r = 100$ and $300$ for the Lorenz systems in chaotic ($r = 28$), intermittently chaotic ($r = 100$), and periodic ($r = 150$) regimes.  $r = 150$ is, in fact, within the period-doubling regime $r \in (145, 166)$. $\epsilon$ and $T_v$ are shown for cW-C in chaotic ($\alpha = 1.3$), quasiperiodic ($\alpha = 1.9$), and periodic ($\alpha = 4.0)$ regimes. For the Lorenz system, RUN outperforms ER in all regimes, although differences are very small for intermittently chaotic dynamics, $r = 100$. In the periodic regime, $r = 150$, $T_v$ reaches the maximum limit set by the observation time $t_o = 109.99$ for $D_r= 300$ and for RUN already for $D_r = 100$. RUN outperforms ER also in forecasting the cW-C system. Only $T_v$ for $\alpha =1.9$ is slightly smaller for RUN than for ER, although $\epsilon$ is smaller for RUN. It is also noteworthy that both RCs predict extremely well also in the periodic regime where dynamics is Hamiltonian, in keeping with findings by Zhang et al.~\cite{zhang21}.

\begin{table}
\centering
\begin{tabular}{|l|c|c|c|c|c|c|} \hline
\multicolumn{3}{|c|}{} & \multicolumn{2}{c|}{$D_r=100$} & \multicolumn{2}{c|}{$D_r=300$}\\ \hline
System & & RC  & $\epsilon$ & $T_v$ & $\epsilon$ & $T_v$\\\hline
Lorenz & $r=28$ & ER  & 0.021 & 5.71 & 0.018 &  6.61\\
 &  & RUN & 0.015 & 6.73 & 0.017 & 6.82\\
 & $r=100$ & ER & 0.123 & 7.31 & 0.089 & 7.47\\
 &  & RUN & 0.120 & 7.37 & 0.125 & 7.53\\
 &  $r=150$ & ER & 0.016 & 77.60 & 0.012 & 109.99 \\
 &  & RUN & 0.016 & 109.99 & 0.017 & 109.99\\
\hline
cW-C & $\alpha=1.3$ & ER  & 0.605 & 13.77 & 0.728 &  16.52\\
 &  & RUN & 0.217 & 25.96 & 0.067 & 39.95\\
 & $\alpha=1.9$ & ER & 0.205 & 31.31 & 0.066 & 105.19 \\
 &  & RUN & 0.090 & 92.41 & 0.053 & 92.41\\
 & $\alpha=4.0$ & ER & 0.628 & 20.27 & 0.663 & 12.19 \\
 &  & RUN & 1.954 & 12.43 & 2.758 & 29.18\\
\hline
\end{tabular}
\caption{Errors $\epsilon$ and valid times $T_v$ measured for $100$ ER and RUN RCs. The Lorenz system in  chaotic ($r = 28$), intermittently chaotic ($r = 100$), and periodic ($r = 150$) regimes, $T_{train} = 1000$. The Wilson-Cowan system in chaotic ($\alpha=1.3$), quasiperiodic ($\alpha=1.9$) and periodic ($\alpha=4.0$) regimes, $T_{train} = 3000$. 
}
\label{tab:LWC}
\end{table}

\subsubsection{Climate replication of the chaotic Lorenz and coupled Wilson-Cowan systems}
\label{climate replication}

To evaluate replication of ergodic properties of the chaotic Lorenz ($r = 28$) and cW-C ($\alpha = 1.3$) systems we measured the spectrum of Lyapunov exponents $\lambda_i$, $i \in [1, 2, \ldots, D_r]$ and correlation dimensions $d_c$ for $20$ optimized RCs of $D_r = 100$ and $300$. We show results only for the three best-performing RCs: ER, RUN and the reservoir including a single cycle (ISC).

\begin{figure}[ht]
  \centering
    \includegraphics[width=0.48\textwidth]{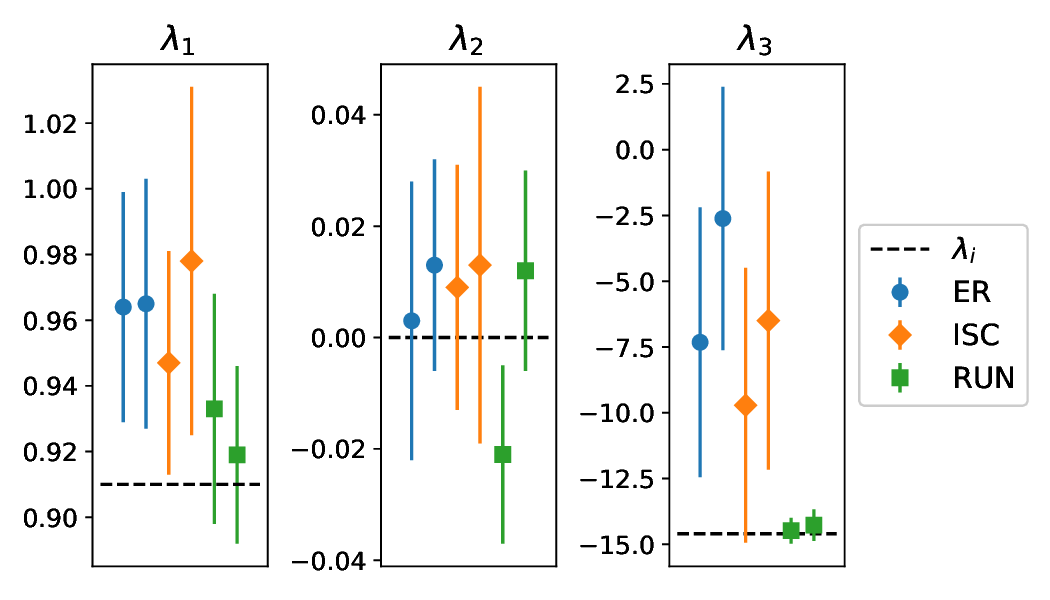}
  \caption{The Lyapunov exponents of the standard chaotic Lorenz system (dashed horizontal lines) and the corresponding estimates with error bars using different reservoirs. Two estimates are shown for each reservoir topology: $D_r=100$ (left) and $D_r=300$ (right).}
  \label{fig:Lorenz_lambdas}
\end{figure}

For RCs replicating the Lorenz system, the first three $\lambda_i$ that correspond to the true dimensions are shown in Fig.~\ref{fig:Lorenz_lambdas}. RUN is seen to give the most accurate estimates. Also standard deviations for RUN are the smallest, which is especially clear for $\lambda_3$. This can be addressed to RUNs having no randomly inserted connections. Also $\lambda_i$ for $i > 4$ are determined best by RUN. Ideally, for the three-dimensional system a RC should give $\lambda_i = -\infty$ for $i > 4$. For each RUN, all $\lambda_i$, where $i \in [4, 5, 6, \ldots 100]$, have a constant very large negative value ranging from $-100$ to $-3500$, whereas for ER and RIS these $\lambda_i$ may be even larger than $\lambda_3$. So, unlike other reservoirs, RUN has very precisely the dimension of the true dynamical system.


Correlation dimension $d_c$ that gives an estimate of the fractal dimension $d_f$ of the signal was computed using Grassberger-Procaccia algorithm~\cite{grassberger_a83, grassberger_b83}. As seen in Table~\ref{tab:Lorenz_dims}, reservoirs give $d_c$ with reasonable precision. Except for ISC, increasing $D_r$ from $100$ to $300$ slightly improves the estimate.
\begin{table} [ht]
\centering
\begin{tabular}{|l|c|c|c|c|} \hline
Lorenz & \multicolumn{4}{c|}{$d_f = 2.06$} \\ \hline
\multicolumn{1}{|c|}{} & \multicolumn{2}{c|}{$d_c$} & \multicolumn{2}{c|}{$d_{KY}$}\\ \hline
 & $D_r = 100$ & $D_r = 300$ & $D_r = 100$ & $D_r = 300$\\ \hline
ER & 2.02 $\pm$ 0.082 & 2.04 $\pm$ 0.067 & 2.225 $\pm$ 0.242 & 2.412 $\pm$ 0.321 \\
ISC & 2.00 $\pm$ 0.067 & 1.98 $\pm$ 0.096 & 2.169 $\pm$ 0.145 & 2.273 $\pm$ 0.301\\
RUN & 2.01 $\pm$ 0.072 & 2.03 $\pm$ 0.094 & 2.063 $\pm$ 0.002 & 2.064 $\pm$ 0.003\\ \hline
\end{tabular}
\caption{\label{tab:Lorenz_dims} Correlation dimension $d_c$ and Kaplan-Yorke dimension $d_{KY}$ estimates of the fractal dimension $d_f$ of the standard chaotic Lorenz system by different reservoirs.}
\end{table}

The upper bound estimate of the signal fractal dimension based on the measured $\lambda_i$, the Kaplan-Yorke dimension $d_{KY}$, see Eq.~(\ref{Kaplan-Yorke}), is computationally more precise than $d_c$.  The computed $d_{KY}$ are given in Table~\ref{tab:Lorenz_dims}. For RUN, $d_{KY}$ is exactly equal to $d_f$ of the Lorenz system. In addition, the error of the estimate is almost two orders of magnitude smaller than for other reservoirs, again due to the absence of randomized links.

\begin{figure}[ht]
  \centering
    \includegraphics[width=0.48\textwidth]{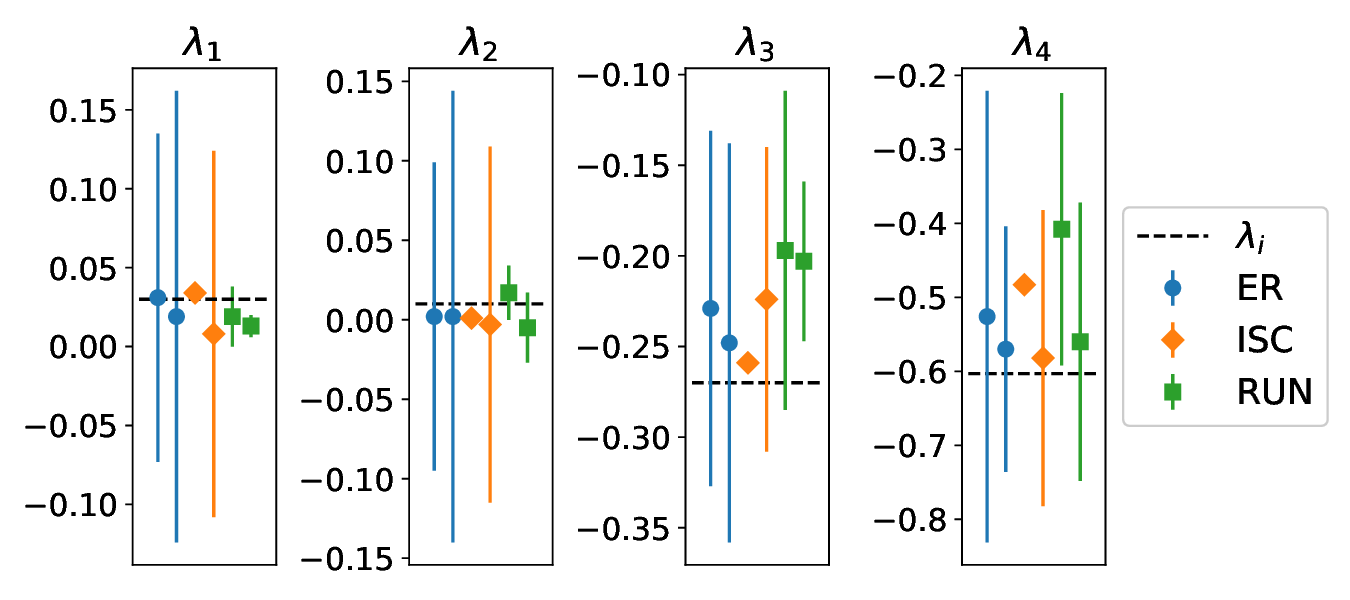}
  \caption{Numerical estimates of the first four Lyapunov exponents of the chaotic coupled Wilson-Cowan system (dashed horizontal lines) and the corresponding estimates  with error bars by 20 optimized RCs of each topology. Two reservoir sizes $D_r=100$ (left) and $D_r=300$ (right) were used for each topology. For clarity, error bars for ISC $D_r=100$ are omitted as the values were considerably larger than for the other cases. }
  \label{fig:cW-C-lambdas}
\end{figure}

The $\lambda_i$ and $d_c$ measured for the coupled Wilson-Cowan system (cW-C) must be compared to numerical estimates of $\lambda_i$ from the data, since there are no exact theoretical values available. These estimates were computed using the classic algorithm introduced by Benettin et al~\cite{benettin80} and Shimada and Nagashima~\cite{Shimada79} and are given together with the estimates by the three RCs in Fig.~\ref{fig:cW-C-lambdas}. Due to the time translation invariance the correct $\lambda_2 = 0$, so the numerical estimation from the simulated data seems to give slightly too large $\lambda_i$. Also errors are very large for ER and ISC, so precise comparison of the performance of the different RCs is not possible. $\lambda_1$ seems to be best estimated by ER. On the whole, $\lambda_i$ are estimated roughly equally well by all RCs as seen from $d_{KY}$ in Table~\ref{tab:cW-C_dims}.

In summary, the climate of low-dimensional chaotic systems is replicated by RUN better than or equally well as by linked RCs.

\begin{table} [ht]
\centering
\begin{tabular}{|l|c|c|c|c|} \hline
cW-C & \multicolumn{4}{c|}{$d_{KY} = 2.12$} \\ \hline
\multicolumn{1}{|c|}{} & \multicolumn{2}{c|}{$d_c$} & \multicolumn{2}{c|}{$d_{KY}$}\\ \hline
 & $D_r = 100$ & $D_r = 300$ & $D_r = 100$ & $D_r = 300$\\ \hline
ER & $1.705 \pm 0.548$ & $1.64 \pm 0.565$ & $2.173 \pm 0.649$ & $2.091 \pm 0.730$ \\
ISC & $1.895 \pm 0.912$ & $1.447 \pm 0.569$ & $2.14 \pm 1.063$ & $2.04 \pm 0.674$\\
RUN & $1.737 \pm 0.295$ & $1.36 \pm 0.207$ & $2.097 \pm 0.418$ & $2.040 \pm 0.410$ \\ \hline
\end{tabular}
\caption{\label{tab:cW-C_dims} Correlation dimension $d_c$ and Kaplan-Yorke dimension $d_{KY}$ estimates of the fractal dimension $d_f$ of the chaotic coupled Wilson-Cowan system by different reservoirs.}
\end{table}

\subsubsection{Summary of forecasting and replication of low-dimensional systems}

The Lorenz and cW-C systems are fundamentally different. The output of the cW-C system in all regimes is extremely spiked compare to the output of the Lorenz system. They have in common that they exhibit different dynamical regimes and are low-dimensional. Only dimension $d_f \lesssim 2.12$  is required to capture the dynamics of the higher dimensional cW-C. The systematic reservoir optimization together with forecasting and replication of simulated data from these low-dimensional systems using sufficient statistics clearly showed that using RCs with linked reservoirs, whether RNNs or just connected networks, did not accomplish the tasks of forecasting or replication as well as RUN, a simple feed forward network.

The advantage of RCs using RNNs in the present context is explained as them being dynamical nonlinear systems. In contrast, a feed forward network like RUN only makes a function transformation of the input to output. The conclusion concerning forecasting and learning low dimensional nonlinear systems, chaotic or not, is that methods based on function transformation and regression outperform traditional reservoir computing. The question then is,  how complex the system under study has to be before the dynamics of linked reservoirs can be used to advantage. This is the topic of the next section.  

\subsection{High-dimensional dynamical system}
\label{high-dim}

Mere functional transformation, which is all that RUN is, was seen to outperform linked reservoirs in forecasting and replication of low-dimensional nonlinear systems. Next, we use a nonlinear system whose dimension can be changed to see if the dynamical aspect involved in RCs with linked reservoirs becomes important in learning increasingly complex systems. A prototype of a system exhibiting high spatiotemporal complexity is the Kuramoto-Sivashinsky (K-S) system, where increasing $L$ in Eq.(\ref{K-S}) transfers the dynamics from the state of traveling waves to chaos.

Table~\ref{tab:KS} shows $\epsilon$ and $T_v$ obtained when forecasting the data from strongly chaotic K-S with $L =35$ for all RC topologies. Considerably larger reservoirs are needed with K-S of $L = 35$ than with the low-dimensional systems. Forecasting is seen to improve for all RCs when $D_r$ is increased from $2000$ to $5000$. We use $D_r = 2000$ that gives representative results to keep the computational cost reasonable. Although RUN is seen to be able to satisfactorily forecast the K-S system with $L = 35$, the linked reservoirs perform better.

\begin{figure}
  \centering
    \includegraphics[width=0.48\textwidth]{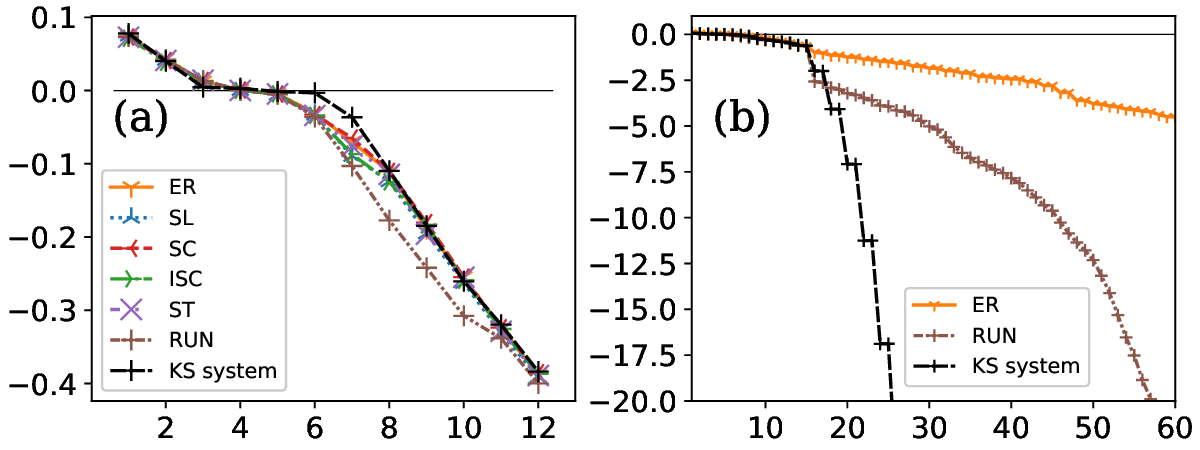}
  \caption{Lyapunov exponents $\lambda_i$ obtained for RCs replicating the Kuramoto-Sivashinsky system, $L=35$. a) $i \in \{1,2,3,...,12\}$  and b) $i \in \{1,2,...,60\}$. All reservoirs fail to reproduce $\lambda_6$ and $\lambda_7$, and $\lambda_i$ where $i \ge 16$. ER succeeds to reproduce $\lambda_i$, $i \in [8, 15]$, where RUN fails.  
  }
  \label{fig:KS_ly}
\end{figure}

Climate replication of this K-S system with $L = 35$ is in keeping with the previous result on forecasting. Fig.~\ref{fig:KS_ly}a) shows the first $12$ $\lambda_i$ obtained over $20$ optimized RCs of each topology together with the numerically computed $\lambda_i$ for the chaotic K-S. RUN gives reasonable estimates, but connected reservoirs perform better.  Fig.~\ref{fig:KS_ly}b) shows $\lambda_i$ for $i \le 60$. It is evident that only the first $15$ $\lambda_i$ are important for replicating the climate of the chaotic K-S system.

It is noteworthy that recurrence is not essential for the RC to forecast or replicate this system exhibiting spatiotemporal chaos, see Table~\ref{tab:KS} and Fig.~\ref{fig:KS_ly}. Only connectedness of the reservoir is of importance. Replication is of equal quality by non-recurrent ST and SL reservoirs and recurrent  ER, SC, and ISC reservoirs.  



\begin{table}
\centering
\begin{tabular}{|l|c|c|c|c|c} \hline
\multicolumn{1}{|c|}{K-S} & \multicolumn{2}{c|}{$D_r=2000$} & \multicolumn{2}{c|}{$D_r=5000$}\\ \hline
 RC  & $\epsilon$ & $T_v$ & $\epsilon$ & $T_v$\\\hline
 ER  & 0.278 & 61.25 & 0.139 &  69.50\\
 ISC & 0.219 & 63.17 & 0.101 & 72.68\\
 ST & 0.229 & 62.82 & 0.116 & 72.30\\
 SC & 0.258 & 62.13 & 0.161 & 67.91 \\
 SL & 0.228 & 63.50 & 0.165 & 67.39\\
RUN & 0.362 & 53.42 & 0.245 & 59.41\\
\hline
\end{tabular}
\caption{Errors $\epsilon$ and valid times $T_v$ measured for $100$ RCs trained for $T_{train} = 5000$ replicating a Kuramoto-Sivashinsky system with $L = 35$.
}
\label{tab:KS}
\end{table}

\begin{table} [ht]
\centering
\begin{tabular}{|l|c|c|c|c|} \hline
 & \multicolumn{4}{c|}{$d_{KY}$} \\ \hline
 & $L = 19$ & $L = 22$ & $L = 35$ & $L = 60$\\ \hline
K-S & $3.47$ & $5.20$ & $7.77$ & $13.56$\\
ER & $3.12 \pm 0.030$ & $4.59 \pm 0.324$ & $7.21 \pm 1.157$ & $13.54 \pm 0.771$ \\
ISC & $3.13 \pm 0.029$ & $4.36 \pm 1.075$ & $7.06 \pm 0.937$ & $13.36 \pm 0.353$\\
ST & $3.10 \pm 0.251$ & $4.31 \pm 0.398$ & $7.14 \pm 0.286 $ & $13.41 \pm 0.550$\\
RUN & $3.12 \pm 0.031$ & $4.29 \pm 0.181$ & $6.86 \pm 0.095$ & $13.71 \pm 0.319$ \\ \hline
\end{tabular}
\caption{\label{tab:K-S_dims} Kaplan-Yorke dimensions $d_{KY}$ of the Kuramoto-Sivashinsky system replicated by different reservoirs. We obtain exactly the same $d_{KY}$ as reported for $L = 22$ and $60$ in~\cite{edson19}.
}
\end{table}



\begin{figure}
  \centering
    \includegraphics[width=0.49\textwidth]{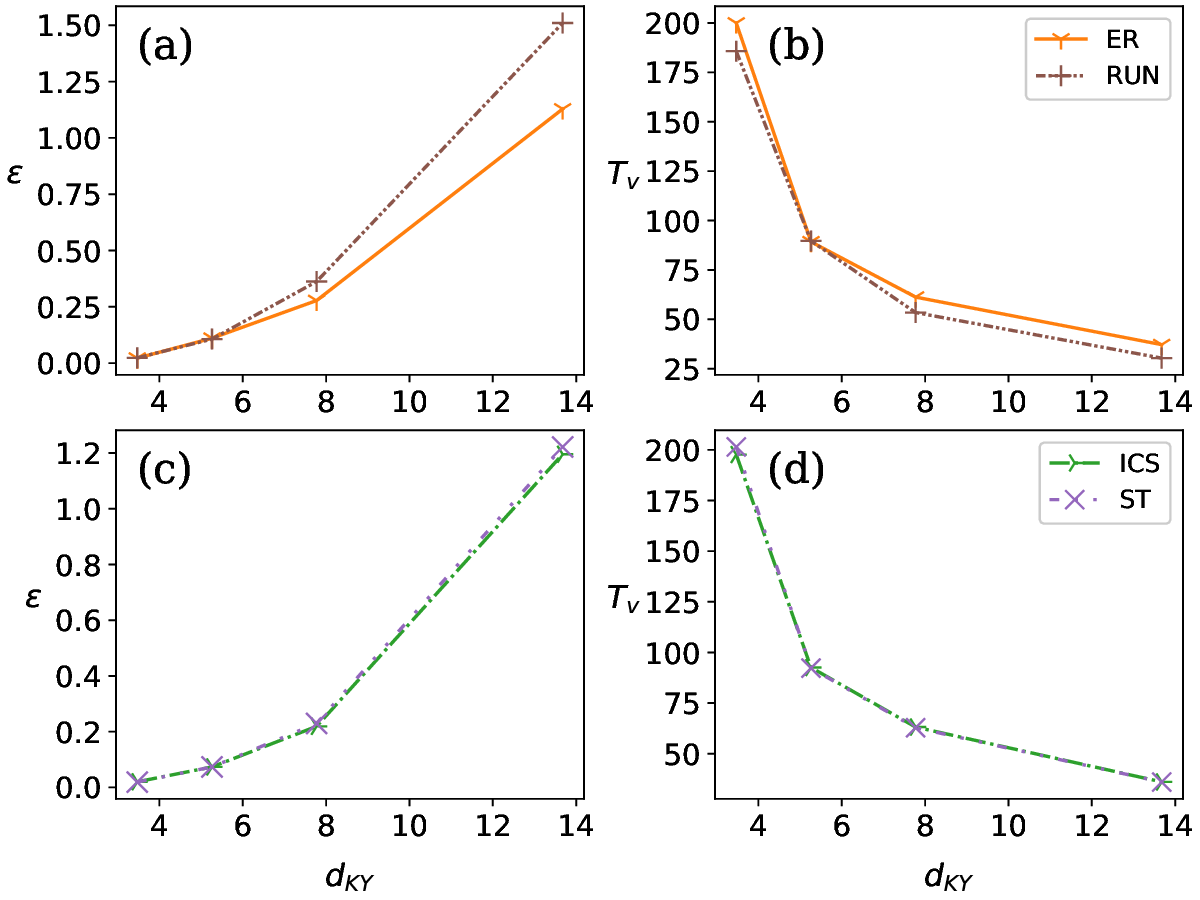}
  \caption{Kuramoto-Sivashinsky systems. a) Error $\epsilon$ and b) valid time $T_v$ vs $d_{KY}$ of $100$ ER and RUN reservoirs generated with optimized parameters. c) $\epsilon$ and d) $T_v$ vs $d_{KY}$ of $100$ ICS and ST reservoirs.}
  \label{fig:KS_astp}
\end{figure}

Comparison of RCs for replicating K-S with different $L$ is made via $d_{KY}$, that is, via the whole spectrum of $\lambda_i$. Table~\ref{tab:K-S_dims} shows $d_{KY}$ obtained numerically for K-S systems of selected $L$ and for three RCs. $d_{KY}$ is seen to be determined for K-S with $L = 19$ and $22$ roughly equally well by all RCs. For K-S with $L = 35$ and $60$, RUN becomes in comparison less accurate.

In Fig.~\ref{fig:KS_astp} we plot the measured $\epsilon$ and $T_v$ vs $d_{KY}$ for ER and RUN. As seen in Table~\ref{tab:K-S_dims} and Fig,~\ref{fig:KS_astp}a), ER performs equally well on K-S systems for which $d_{KY} \lesssim 5.5$. For $d_{KY} > 5.5$, $\epsilon$ increases with increasing $d_{KY}$ at a higher rate for RUN than for ER. The qualitatively similar result obtained for $T_v$ is plotted in Fig.~\ref{fig:KS_astp}b).

Figs.~\ref{fig:KS_astp}c) and d) show $\epsilon$ and $T_v$ vs $d_{KY}$ for ISC and ST. In keeping with our findings for replication, it is seen that recurrence of ISC does not improve forecasting of the K-S system from what is already achieved by using the connected, non-recurrent ST.

The role of ESP is confirmed by results on RUN learning the K-S dynamics. Leakage is minimal or non-existent in all RUNS optimized and used for K-S. The median values from optimizations are: $\beta = 0.71,\ 0.8,\ 1$, and $1$ for $L = 19,\ 22,\ 35$, and $60$, respectively. So, for $L = 35$ and $60$ there is no leakage. In RUNs leakage is the only means by which ESP can be realized. Introducing leakage by changing $\beta$ from $1$ to $0.8$ for $L = 35$ did not change $\epsilon$ or $T_v$ in Table~\ref{tab:KS}. This confirms that ESP, originally introduced as a means for achieving convergence with RNNs is exactly that. It is not a prerequisite for a RC to work in general. 


\section{Conclusion}
\label{conclusion}

According to the maximum-entropy principle the information should be obtained from the probability distribution that maximizes the entropy, subject to the constraints~\cite{jaynes57}. The last part on constraints ensures that this fundamental principal of information theory is always valid. It is via these constraints that extracting maximal information can be made more efficient. In the present context this would mean, for instance, benefiting from a linked reservoir being a dynamical system that can be made to mimic dynamics of the target system even over a short time interval. RNN being a highly nonlinear dynamical system is a common argument for using RCs instead of conventional methods for learning chaotic systems.

The maximum-entropy principle in the context of RCs states that the entropy of the system consisting of the reservoir and the output matrix $\mathbf{W}_{out}$ should be maximized subject to the information, which constitutes samples $\mathbf{u}(t_k)$ at times $t_k$, $k \in [1, n-1]$, before the present time $t_n$. This way the entropy of $\mathbf{u}(t_k)$ will be minimized. As each node is connected to the output via $\mathbf{W}_{out}$, the probability distribution, whose entropy is to be maximized, is determined by the reservoir. We form the predicted signal value as $u_i(t + \Delta t) = \sum_{k=1}^M c_k f(r_k(t))$, where $c_k$ are the elements of the matrix $\mathbf{W}_{out}$ and $f(\cdot)$ includes the nonlinear activation, here $\tanh(\cdot)$, and the symmetry-breaking at the output. For $j \ne i$, $r_i(t)$ are determined independently of $r_j(t-\Delta t)$ by regressive fitting of $c_k$ in RUN, whereas in linked networks node $i$ may be connected to node $j$ and, accordingly, $r_j(t)$ depends directly on $r_i(t-\Delta t)$. The probability distribution of the reservoir can be written as $\sum_{k = 1}^M p(r_k(t))$, where $p(r_k(t))$ is the probability of node $k$ having value $r_k(t)$ at time $t$. The entropy of this distribution is at maximum when there are no connections between nodes, since connecting node $i$ to $j$ induces direct dependence of $p(r_j(t))$ on $p(r_i(t-\Delta t))$, that is, connecting node $i$ to $j$ means that $p(r_j(t))|r_i(t-\Delta t)) > 0$ and the corresponding conditional entropy is smaller than the entropy $p(r_j(t))$ of the independent node $i$. The reservoir entropy is formed by constituent entropies of each node. Hence, in the absence of additional constraints, the entropy of any connected reservoir is smaller than the entropy of RUN.

The predicted signal $u_i(t + \Delta t) = \sum_{k=1}^M c_k f(r_k(t))$ is presented with functions $f(r_k(t))$.  In RUN, $r_k = \sum_i b_i u_i(t)$, when there is no leakage. In linked reservoirs $r_k = \sum_i b_i u_i(t) + \sum_l g_l r_l(t-\Delta t)$, where $i$ runs over a random selection of components of $\mathbf{u}$ and $l$ runs over indices of nodes that are linked to node $k$. $b_i$ and $g_l$ are coefficients. In RUN, in the absence of leakage, $u_i(t + \Delta t) = \sum_{k=1}^M c_k f(r_k(t))$ is a mere functional transformation. No dynamical aspect is involved. In contrast, in linked reservoirs the delayed signals, proportional to $r_l(t-\Delta t)$ from nodes $l$ connected to node $k$, are summed at node $k$. In RNNs there may also be terms proportional to delayed signals from the node itself, $\propto r_k(t-(n+1)\Delta t)$, where $n$ is the number of nodes in the loop starting and ending at node $k$. The delayed signals make all linked reservoirs dynamical systems. In other words, there is a dynamical response to the input, instead of just a functional transformation of the input as in RUN. 

No reservoir of any topology produces nonlinear terms by itself. Only linear terms $\propto r(t)$ are summed in the nodes. Nonlinear terms that aid in regression of highly nonlinear signals are due to $f(\cdot)$ only. In other words, no reservoir has an advantage over others because of possessing terms of some form lacking in others. This fact combined with what was shown above means that the dynamical aspect of the linked reservoirs, as opposed to the mere functional transformation in RUNs, is the only difference between these reservoirs that affects supervised learning. 

The dynamical aspect of RCs should be the more important the more complex the dynamics of the system to be learned is. This was seen in our study. RUN performed best in learning systems whose chaotic dynamics on the attractor can be described by approximately $5.5$ modes or less, as given by the measured $d_{KY}$. RCs with linked reservoirs outperformed the RUN in predicting and replicating chaotic systems the description of whose dynamics required more than $5.5$ modes. Recurrence of the reservoir did not enhance forecasting or replication of systems whose chaotic dynamics involves up to approximately $13.5$ modes. RNNs and linked RCs without loops performed equally well. It would seem plausible that there be another transition at a higher value of $d_{KY}$ beyond which RNNs outperform non-recurrent linked networks. Determining this value would require optimization and simulation of even larger reservoirs than used here and is beyond the scope of the present study.

Lastly, the present study shows that obtaining statistically significant results requires judicious hyperparameter optimization and sufficient statistics. Machine learning (ML) is an approach that is more computation intensive than computer simulation of a physical system based directly on dynamical equations. Reservoir computing, although a simplified ML method, is no exception to this rule. Still, the extra effort for optimized RCs and sufficient statistics needs to be made in order to obtain valid results that aid in gaining insight on this rather modestly understood method.

\section*{Acknowledgement}
We acknowledge the computational resources provided by the Aalto Science-IT project.

\newpage
\begin{figure}
\centering
\includegraphics[width=0.70\textwidth]{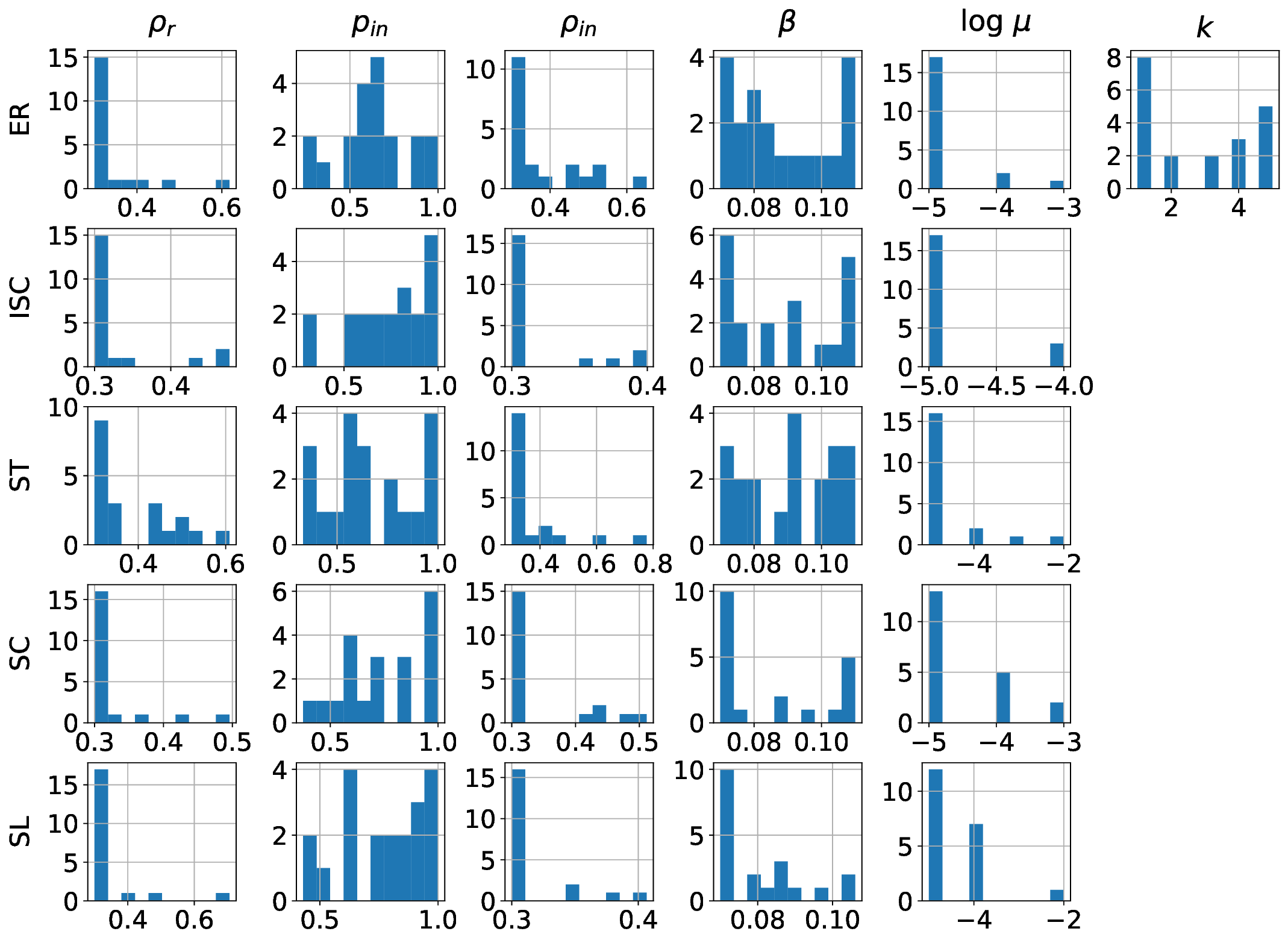}
\caption{Histograms of optimized parameter values for each topology. Topologies from top down are an ER network where $k \in [1, 5]$, ISC, ST, SC, and SL. Optimization ranges R (see Table I). Distributions were found to be quite similar when optimization was run for different sets of $20$ reservoirs.}
\label{fig:paramhist_}
\end{figure}

\end{document}